# Generative AI, Pragmatics, and Authenticity in Second Language Learning

Robert Godwin-Jones, Virginia Commonwealth University

**Abstract**

There are obvious benefits to integrating generative AI (artificial intelligence) into language learning and teaching. Those include using AI as a language tutor, creating learning materials, or assessing learner output. However, due to how AI systems understand human language, based on a mathematical model using statistical probability, they lack the lived experience to be able to use language with the same social awareness as humans. Additionally, there are built-in linguistic and cultural biases based on their training data which is mostly in English and predominantly from Western sources. Those facts limit AI suitability for some language learning interactions. Studies have clearly shown that systems such as ChatGPT often do not produce language that is pragmatically appropriate. The lack of linguistic and cultural authenticity has important implications for how AI is integrated into second language acquisition as well as in instruction targeting development of intercultural communication competence.

**Introduction**

Generative AI (artificial intelligence), as represented by ChatGPT and other systems, offers significant affordances for language learning and teaching. Second language learners are using AI chatbots to practice speaking or to improve their writing. AI can take on the role of language tutor, providing feedback and learning materials. Teachers can use AI to create level-appropriate stories/exercises, help assess student work, or even plan out a curricular unit. A flurry of blog posts, conference papers, and peer-reviewed articles have explored the affordances of AI for language learning and teaching (Godwin-Jones et al., 2024; Kohnke et al., 2023; Poole & Polio, 2023). We are still in the early, exploratory stage of seeing how AI can enhance the language learning process, but it is evident already that AI is being widely used by both learners and teachers and is likely to bring about significant changes in language learning, teaching, and assessment (Godwin-Jones, 2024).

AI users of all kinds have marveled at the proficiency and speed with which such systems produce texts that seem indistinguishable from human writing: coherent, well-structured, and grammatically flawless. AI's language abilities come from machine learning applied to a vast language dataset, building a "large language model" (LLM) which uses statistical probability to predict the next likely word in a text sequence. That ability is then extrapolated to generate phrases, sentences, paragraphs, complete discourses. This mathematical model of language in AI has no lived experience with humans or human society, limiting its understanding of the social and cultural dimensions of human interactions and of human language use. Furthermore, the dataset used to build an LLM is not representative of the world at large, as it draws from digital sources that are overwhelmingly in English and reflect the values and behaviors of Western, industrialized cultures (Naous et al., 2023). This has serious consequences for the social and cultural dimensions of AI output, calling into question its linguistic and cultural authenticity. In particular, AI is incapable of handling the nuances of pragmatic language use, as that requires an ability to negotiate sociocultural common ground with the interlocutor. That has implications for the viability of AI chatbots as effective vehicles for second language learning, especially at more advanced proficiency levels.

Most studies of pragmatics in AI use the Grice cooperative principle (Grice, 1989) to examine whether AI output follows the Gricean maxims of quality, quantity, relevance, and manner (Harnad, 2024; Sydorenko et al., 2024); others have looked at performance related to speech acts (Gubelman, 2024; Tao et al., 2024) and implied meanings (Ruis et al. 2024). Those studies point to severe deficiencies in AI's pragmatic abilities. A useful model for understanding those limitations are studies of intercultural pragmatics (Kecskes, 2013, 2015) as the lack of *a priori* common ground (linguistic and cultural) between speakers with different first languages (L1) mirrors the human-AI relationship. While AI output is often problematic in its linguistic and cultural authenticity, it can still serve useful roles in L2 learning and teaching. In particular, interactional or functional authenticity can be achieved through leveraging AI's writing assistance (especially its genre awareness) to have learners engage in real-world writing tasks.

**Generative AI's Language Abilities**

Artificial intelligence itself is not new, but generative AI represents a sharp break from earlier efforts. While AI was initially designed to function as an "expert system" within a narrow range of functionality, ChatGPT and other generative AI are general use systems that work within a broad range of contexts and can generate language in a wide variety of genres. Additionally, natural language processing was developed in traditional AI by training systems in the rules of how language works (i.e., syntax, morphology, etc.) as well as programming them with knowledge about the social and natural world in which humans live (something akin to humans' common sense). Early AI systems met with only limited success. It turns out to be too complex and vast a task to program computers with enough rules and facts to deal adequately with phenomena as complex as human language and human society (Sutton, 2019).

A different approach to natural language understanding and processing has yielded incredibly better results, namely that used in generative AI. The LLMs that power generative AI systems are created by feeding the system a huge amount of data (digital texts scraped from online sources), turning the text chunks into mathematical symbols ("vectors"). Then, deep machine learning (multiple neural networks operating simultaneously) enables the system to uncover and store representations of patterns and regularities ("parameters"). That results in a statistical, not a linguistic model of language. This is very different from how humans acquire language, through a gradual process of socialization. The statistical model of language in AI is very effective at generating output that seems in its fluency, flow, and idiomaticity to have come from a human source. While AI output seems "authentic," and is certainly grammatically correct, studies have shown that the texts it produces tend to be bland and uninspired (Kramsch, 2023; Poole & Polio, 2023). They lack the spark of human creativity, representing as they do simply repackaging of their textual training data. AI output is form without substance.

AI has no real understanding of the texts it generates and no lived experience in our world to go by to judge the appropriateness of its output for the human user with whom it is interacting. Those limitations result in the fact that AI can "hallucinate," producing statements that are false or inappropriate. Hallucination is not actually an accurate term to use, in that it implies AI is fundamentally in a normal state of reality from which it occasionally and temporarily departs. In fact, LLM's are totally untethered from reality, relying on internal processes based on their training data. AI systems may provide false information, but they do not lie or hallucinate, as that would imply that they have an awareness of truth or an intent to deceive. They are designed to give the impression of objectivity and accuracy but there is no active, intentional being behind the output (Hicks et al., 2024). As Giannakidou and Mari (2024) comment, AI is "epistemologically empty" (p. 3), or, as Shanahan et al. (2023) put it, with AI there is "no-one home" (p. 7).

Studies have shown that, unsurprisingly, AI systems lack cultural sensitivity (Cao et al., 2023). That derives in part from their training data, that represents the largely Western, "first world", male-oriented cultural orientation of the Internet, its principal training source (see Atari et al., 2023). That has been shown in a number of studies that examine AI output from the perspective of particular cultures. Naous et al. (2023) show how Western biases are evident in soliciting Arabic output from an AI. Testing 16 different LLMs, the researchers found consistently cultural faux-pas and inaccuracies, even in Arabic-trained systems. When asked to provide a completion task based on suggested activities after Islamic prayer, an AI listed as one option engaging in drinking alcohol ("going out for a drink"), clearly an activity incongruous with the given cultural context. In storytelling output, the LLM revealed cultural stereotypes such as associating Arabic names with poverty while positive attributes were associated with Western names. Cultural surveys in the contexts of Asian (Wang et al., 2023) and East European (Masoud et al., 2023) contexts found that output reflected Western values, such as attitudes towards individualism. Stokel-Walker (2024) points out that AI systems "think" in English even when using other languages; in other words, they adopt norms and values associated with Anglophone cultures. That makes it problematic for the quality of generated content for languages other than English, especially for less commonly taught languages (Poole &Polio, 2023).

It is likely that casual users of ChatGPT and other systems are unaware of the built-in cultural and linguistic biases. The high quality of the output—and the speed at which it is delivered—tend to convey the impression of an authoritative voice. There is no hint of hesitation or uncertainty on the part of the AI system. An unsuspecting user could be excused for thinking that the generated texts are objective and trustworthy. Indeed, researchers examining the use of AI in second language acquisition (SLA) have not always recognized the problematic nature of LLM output. Some have embraced AI output as an option for "authentic" language use. Chiu et al., (2023) suggest that AI chatbots can "facilitate an authentic, interactional language learning environment" (p. 2), while Kohnke et al. (2023) state that "ChatGPT supports language learning by simulating authentic interactions" (p. 3). Anis (2023) claims AI use is "essential for promoting cultural sensitivity, intercultural competency, and global awareness" (p. 64). AI-based VR has been seen to supply "the benefits of in-country immersion programs without the hassle" (Divekar et al., 2022, p. 2354). The suggestion in these studies is that AI output is linguistically and culturally authentic enough that it could substitute in language learning settings for human interlocutors or could even provide similar benefits to a study abroad experience.

Such a view ignores the process used by AI systems to reproduce language and the limitations of that process for the linguistic features and cultural content of the resulting output. As described above, AI systems break down human language into mathematical symbols to construct a model for language understanding and production. Humans, in contrast, are socialized into their language abilities, learning gradually how to use language appropriately within an ever-expanding set of social circles. In the process, we acquire the ability to adapt to the communicative context, adjusting our utterances, our nonverbal signaling, and our paralanguage (tone of voice, pitch, volume, etc.) to fit the situation and our conversation partner: that is the nature of an *authentic* human conversation. The different ways language is learned and used by AI and by humans is central to understanding the nature of AI output and why, despite its considerable language abilities, it cannot fulfill the role of conversation partner in the same way a human being can. It is very good at mimicking human conversation, even simulating interest and concern. AI training data contains enough emotional language (from novels, dramas, movie transcripts) for the system to draw on those forms and expressions. No emotions or feelings are present, of course, although, once again, AI's mimicry can seem believable, particularly as humans have the tendency to anthropomorphize machines that seem responsive (Holtgraves et al., 2007). That is all the more the case with a machine as glib with language as an LLM.

**AI, Authenticity, and Pragmatics**

The concept of authenticity in language learning is complex and has proven to be controversial (2013, Buendgens-Kosten, 2014; Mishan, 2005). Gilmore (2019) provides seven different kinds of authenticity in language learning. How authenticity is understood and valued is linked to the dominant pedagogical approach in effect at the time (Zyzik & Polio, 2017), so that it has become something of a moving target. There is not unanimity among SLA research that authenticity is even desirable (Long, 2014). Furthermore, how authentic texts or tasks are received and used by learners is unpredictable in that "L2 developmental trajectories are likely to be highly idiosyncratic" (Gilmore, 2019, p. 305). No matter what approach is used, not all variables in language learning can be controlled, so that what each participant gets out of an "authentic" L2 encounter may be quite different. Gilmore (2019) concludes his examination of different versions of authenticity in asserting that a social constructivist perspective on authenticity is closest to the lived experience of language teachers. That interpretation of the term stresses the social connectivity of a L2 text or task. That view accords with other scholars who have highlighted the interactional aspects of authenticity, emphasizing the need in designing learning activities to reflect the original communicative purpose of the task, to approximate real life interactions, and to involve purposeful communication (Mishan, 2017; Poole & Polio, 2023). If we are to consider authenticity from the perspective of interactions, then that brings into play the social and cultural dimensions of communication, essential for effective conversational exchanges. That places a greater weight on language pragmatics.

According to the widely used cooperative principle in pragmatics (Grice, 1989), conversation in normal social settings is guided by a desire to cooperate and accommodate ones partner. Cooperation in conversation necessitates on-the-fly adaptation based on reading the interlocutor's state of mind, emotional response, and physical reactions. This is clearly problematic for AI. The statistical model of language in AI lacks the sociocultural grounding humans have through sensorimotor interactions and from simply living in the real world. Studies of AI's capabilities to engage in pragmatically effective language use have shown significant limitations (Lee & Cook, 2024; Su & Goslar, 2023; Tao et al., 2024). While AI systems can gain pragmalinguistic knowledge and learn appropriate formulaic sequences (conventions for apologizing, for example) through the verbal exchanges in their training data, they have proven to be much less effective in sociopragmatic engagement, that is, in generating contextually acceptable speech reflecting an interlocutor's state of mind, intentions, and emotional status. Pragmatic appropriateness is not pre-set but is developed in interaction. Pragmatic meanings "do not inhere in linguistic conventions but result from participants' ongoing, contingent interpretive work during jointly pursued practical activities" (Kasper, 2009, pp. 278–279). Knowledge of normalized formulae is not enough. Contextual factors such as social distance, relative power, ranking of imposition determine pragmatic language, as do concerns such as face, rights, and obligations (Sykes & González-Lloret, 2020). There is a complex sociocultural process at play.

A study by Gerhalter (2024) reflects others that have examined the pragmatic competence of AI systems. That paper examined the use of topicalized infinitives in Spanish and Portuguese, with the finding that ChatGPT had no trouble with understanding meanings but could not properly integrate their use in conversational exchanges nor interpret inferred meanings they contained. The author concluded: "ChatGPT is good at mimicking common speech patterns, but has no idea what it is talking about" (p. 32). As every L2 learner has experienced, knowing the appropriate forms and formulae is not enough for communicating effectively; those constructions need to be integrated properly into the give and take of a conversation. That process will require interactional and strategic competence that comes from human experiences in carrying out speech acts such as greetings or apologies, negotiating cultural/personal characteristics such as degree of directness or shyness, or learning how to maintain a conversation (topic selection/change, repair, role of silence, closings). Dombi et al. (2024) comment that for developing such conversational skills, synthetic conversations will not suffice; human conversations are needed for "allowing learners to practice turn-taking and social language using natural discourse patterns" (p. 43). In addition to sociopragmatic limitations, studies of AI and pragmatics have shown that in some cases, ChatGPT may even struggle with the pragmalinguistic forms themselves, as Lee and Wang (2023) found for the use of politeness conventions in Korean (using honorifics, recognizing hierarchical distinctions). It is likely that performance in pragmatics will vary considerably across languages, due to the skewed linguistic training data which heavily relies on English.

More studies of the pragmatic language abilities of ChatGPT have recently emerged. Several use the well-known Gricean maxims for guiding effective communication in social situations, supporting Grice's cooperative principle (Grice, 1989). Giannakidou and Mari (2024) found that ChatGPT violates the Gricean principle of Quality, which demands that speakers be truthful. Because AI systems have no real-world understanding of the texts they produce, they are prone to inaccuracies and even false information. That possibility also can lead to violation of the Gricean maxim of Relevance (all statements should be relevant to the conversation; see Harnad, 2024). Also suspect is the maxim of Manner (be clear and straightforward), as ChatGPT is often overly polite and formal (Chen et al., 2024; Dynel, 2023). Studies have shown how ChatGPT also violates the Gricean principle of Quantity (give as much information as needed, and no more) as the output is often overly verbose and repetitive (Barattieri di San Pietro et al, 2023; Harnad, 2024).

ChatGPT has been shown to have difficulty resolving implied meanings (Barattieri di San Pietro, 2023), a major topic in studying pragmatics. Implicature is an essential component of meaning-making in human conversation. In the study by Ruis et al. (2024), a variety of AI models were tested and the overall accuracy in resolving implied meanings was close to a random level (around 60%), while humans obtained 86% accuracy. The study found that providing fine tuning with explicated examples or using chain of thought prompting techniques enhanced considerably the performance of the AI systems. Other studies have shown as well that additional training improves pragmatic abilities. Chen et al. (2024) used discourse completions tasks and role plays to test ChatGPT and found that in most instances, the AI performed as well as humans in both pragmalinguistic and sociopragmatic areas. However, the results were obtained by providing to the AI systems "attitudinal indicators" that were not included in the scenarios used by the human subjects. After presenting a role play testing performance with apologies, for example, the study provided this prompt to ChatGPT: "Write the conversation as if you feel apologetic and as if the classmate's friend feels annoyed" (p. 17). In other words, the AI was instructed how to "feel" so that appropriate language could be generated. Similarly, in a study using implied meanings in short dialogues, AI needed added hints to uncover the real meaning in the implicature (Brunet-Gouet et al., 2023). The fact that AI chatbots need help to respond appropriately in speech acts or in other pragmatically informed contexts, has implications for AI use as free conversation practice partners in L2 learning and teaching. On the other hand, the possibility of adding additional pragmatics training to an AI system would allow for building a custom GPT specifically designed for use in SLA (see Lan & Chen, 2024), not a solution available to all L2 learners.

Learning to respond appropriately to an L2 conversation partner is a critical part of developing interactional and pragmatic language competence. As discussed above, that is problematic for AI chatbots. While the LLM may supply the necessary pragmalinguistic forms to use in a given context, how, when, and if that language is used depends on the human partner and her personal situation. There are a number of factors at play. One is the fact that in any given social interaction, there are a variety of possible behaviors and linguistic responses: "Casual conversation has a wide range of allowable and unpredictable contributions which are negotiated in the flow of communication" (Bardovi-Harlig, 2020, p. 55). Some of the variations are culturally determined and others highly personal. An L2 learner, for example, may for personal, interpersonal, or sociopolitical reasons modify, or outright reject expected pragmatic choices. McConachy (2019) gives an example of a learner from an egalitarian society feeling discomfort in adapting to linguistic expectations in a hierarchally oriented culture. Learners may feel most comfortable adopting the role of cultural outsider, thus not participating fully in native-normalized pragmatic behaviors (García-Pastor, 2020). Using AI for learning pragmatic norms might run counter to individual identity positioning, given that it will assume the mainstream (WEIRD) values and behaviors it has learned from its training data (Atari et al., 2023). Akane et al. (2024) illustrate that potential. The authors advocate using AI output to demonstrate to Japanese learners of Chinese to be more "direct" in their conversational strategies, in other words, to be more like Westerners.

Another factor that constrains effective pragmatic language use in AI is the assumption that one can rely on the sociopragmatic norms of a single homogeneous speech community. This is in fact, as Bardovi-Harlig (2020) points out, a failing often seen in instructional pragmatics as well as in textbooks. In reality, the norms for any language vary according to the nature of communities, whether they be geographically or linguistically defined. In judging pragmatic appropriateness, one would need to take into consideration local characteristics including language varieties such as dialect. That is especially the case for pluricentric languages like English, Spanish, or Arabic (Bardovi-Harlig, 2020). McConachy (2019) argues convincingly that in considering pragmatics, "we need to move beyond a view of pragmatic awareness that is built upon a view of language use as a tight normative system centred primarily on form-function-context mappings which are evaluated with reference to the central criterion of 'appropriateness'" (p. 173). He argues that pragmatics is not just a form of social action but also a form of moral action and that language used by individuals is interpreted through personal and cultural "schemas for rules, rights, and obligations and interpersonal relations" (p. 170) developed through previous interactions.

**Intercultural pragmatics, cultures-of-use, and AI**

To explore the issues that AI systems have in dealing adequately with pragmatically rich contexts, it can be helpful to look to intercultural pragmatics (Kecskes, 2013, 2015, 2019; McConachy, 2019). Gricean approaches to pragmatics tend to focus mostly on English in monolingual contexts (Kecskes, 2014). Given the reality of multilingualism in most parts of the world today (especially in online environments), pragmatics needs to be able to explain behaviors and language use in multilingual and multicultural settings (Kecskes, 2015). The Gricean cooperative principle assumes that conversants in social situations are rational, well-meaning, and intentional in striving to achieve a mutually beneficial outcome (Tao et al., 2024). That process presumes no ambiguity, dissembling, or misrepresentation on the part of speakers (Setlur & Tory, 2022), which may not universally be the case.

Gricean pragmatics assumes that speakers have the same goals and are able to use common ground to facilitate effective communication. That common ground is built on accepted cultural norms and behaviors drawn from prior life experiences and shared linguistic practices, such as the use of familiar formulaic expressions (Dombi et al., 2022). In intercultural encounters and in L2 conversations with speakers of different L1s (in English as a lingua franca settings, for example), the common ground based on shared culture and language is missing. Instead, an emergent and reciprocally constructed common ground needs to be created by the conversation partners, negotiated on the fly (Dombi et al., 2022; Kecskes, 2015). Studies of English as a lingua franca have shown how non-native English speakers in conversation are characterized by a mutual conscious strategy to cooperate (Baker & Sangiamchit, 2019; Godwin-Jones, 2020). That is different from the non-reflective, automatic linguistic and cultural common ground among common L1 speakers (Dombi et al., 2024). In intercultural contexts, common ground is achieved through a process of adapting language to the evolving conversation and to the reading of the other person's state of mind and feelings. In that setting, speakers need to overcome the "communication asymmetry" (Dombi et al., 2022, p. 8) based on the absence of a shared system of knowledge and experience (Balaman & Sert, 2017).

Finding common ground in intercultural encounters increasingly occurs in online settings. That fact adds another layer to the negotiation of identity positions and linguistic/cultural commonalities. Technologies have their own "cultures-of-use" (Thorne, 2003), pragmatic norms for acceptable user activity. Digital tools and services are not neutral but themselves can have a co-determining influence on how a communicative event plays out (Darvin, 2023). Thorne (2003) found that there were very different norms and expectations for how exchanges between learners with different L1s took place in email, online forums, or different forms of computer-assisted communication. With technology use, there is in effect a kind of double intercultural layer:

> Virtual communication creates its own cultures and expectations, opening spaces for double intercultural communication, where not only different language and cultural backgrounds come into contact through the task, but also new, unexpected, digital cultures may emerge on the fly through the technology (González-Lloret & Ortega, 2018, p. 208).

The kind of "small culture" created in online communities or in virtual exchange (Holliday, 1999) is the result of interlocutors engaging in a process of give and take and probing in terms of factors such as tech tool familiarity, L2 proficiency, personal preferences, and cultural backgrounds.

The added complexity technology plays in human-to-human exchanges is clearly in evidence in the communicative process as well between humans and AI. With AI, the technology tool itself becomes the conversation partner. As in intercultural encounters, there is no common ground, as AI has no life experiences to share. Similar to intercultural encounters, AI users engage in back-and-forth interactions, testing the system through trying out different prompts to see how the AI responds. The negotiation of common ground is guided by the user, not co-constructed as in human-to-human conversations. Studies have shown that in chats with AI, humans tend to use similar strategies in guiding the conversations as are used in intercultural or English as a lingua franca encounters (Dombi et al., 2022; Kramsch, 2023). That process may involve significant modifications to language use. Dombi et al. (2022) has shown that in computer interactions humans tend to use shorter sentences, avoid figurative language, and utilize more careful word selection (see also Timpe-Laughlin et al., 2022). Social language (small talk) is kept to a minimum and behaviors such as self-repair, repetition, and elaboration are avoided. Suprasegmental features may differ as well (loudness, intonation, rhythm). In conversations with a computer there is more conscious recipient design than would be the case in human-to-human conversations (Sydorenko et al, 2024). There is a continual monitoring of language use with, as in intercultural exchanges, more use of semantically transparent language rather than idiomatic expressions (Kecskes, 2015).

The linguistic adaptations humans tend to use in interactions with AI have an impact on the usefulness of AI chatbots for language learning. Clearly, chatbots can serve important roles in different stages and in different contexts of second language acquisition. At novice levels–and particularly in the absence of available L2 conversation partners—chatbots represent a valuable resource for conversation and pronunciation practice. They can serve as a kind of native informant, providing assistance and feedback on demand. AI exchanges can build confidence and enhance motivation. Their constant availability (including on personal devices) makes them a constantly available L2 partner. At the same time, even beginner-level learners should be aware that AI chats are not equivalent to conversing with a fellow human. As we engage more with AI agents, both teachers and learners will "need to know what exactly students are practicing in such technology-mediated tasks without 'real' human interlocutors in order to utilize the technology most effectively for L2 practice" (Dombi et al., 2024, p. 35). We need critical AI literacy for all educational stakeholders to appreciate the contexts in which AI use can benefit learners and others in which AI will not provide the needed L2 experience. While AI chats can be useful in practicing transactional language, they are less so when it comes to the ability to use language to develop harmonious social relationships (Dombi et al., 2024).

The remarkable language capabilities of AI systems that mimic high performing human speakers or writers may lead users to accept AI output as equivalent to what a human in a conversation might say. But, as we have seen, AI language is not "natural" or "authentic" in the same way that humans communicate through leveraging their social and cultural awareness to adapt to a conversation partner, whether that be someone of a shared culture or not. The common ground is easier to achieve for common L1 speakers, but with conscious effort and good will we manage in intercultural contexts as well. For non-novice learners, the tendency in conversing with AI to avoid idiomatic and figurative language marks the conversation as unnatural. If pragmatics is "at the core" of language (Syderenko, 2020, p. 48), then formulaic language is the "heart and soul of native-like language use" (Kecskes, 2015, p. 21). In intercultural settings, formulaic language is reduced in favor of more immediately understandable formulations (Kecskes, 2015). That lessens the usefulness of chatbots for developing native-like language abilities.

While AI may not be a fully satisfactory partner when it comes to pragmatic language use, that very failing could prepare L2 learners for intercultural encounters. In English as a lingua franca and other similar intercultural settings, speakers need to be able to communicate in many varied contexts with speakers from a variety of backgrounds (Sydorenko et al., 2024). Encounters with AI, lacking as they are in common ground, could prepare L2 learners for that experience "by practicing adapting to an ongoing, potentially less natural discourse" (Dombi et al., 2024, p. 47). Learners could learn the importance of being flexible and adapt their turns and language relative to the needs of their interlocutor and to the flow of the conversation.

**AI and Functional Authenticity**

Poole and Polio (2023) point out that although, as we have seen, AI output lacks both linguistic and cultural authenticity, its use can be *functionally authentic* through using AI to help produce texts used in settings beyond the classroom. Authenticity related to interactions in extramural settings has increasingly been seen as more critical in instructed SLA then looking at the authenticity of texts or tasks in isolation (Gilmore, 2019; Mishan, 2015). Poole and Polio (2023) present examples that feature real-world integration such as writing a restaurant review for Yelp or generating tweets for marketing a product. Similar "renewable assignments" that feature real-world integration of student writing are outlined in Blyth (2023) and in the studies discussed in the collection by Ryan and Kautzman (2022). In the process of writing for real audiences, students see that AI literacy is something useful beyond the classroom. Comparing real-world texts with AI generated content can be revelatory in terms of what AI can do well–writing high quality connected discourse–and where it might be deficient—understanding and integrating social and cultural nuance. In shared writing with humans, AI functions as a mediating tool, as does, crucially, the teacher in guiding learners towards critical evaluation of AI output.

One approach to exploring AI use in functionally authentic tasks is to generate examples of particular genres of writing. That could range from simple, everyday text forms (blog posts, Amazon reviews, YouTube comments) to professional/academic genres (news reports, abstracts, peer reviews) to multimodal examples (social media). Poole and Polio (2023) point out that for some genres authentic model texts may not be easily available, such as for letters of recommendation. In such cases AI could provide helpful models. The authors provide the example of an email in an academic setting. A student has missed a quiz because of illness and needs to explain the situation to the instructor. A variety of classroom tasks could be based on comparing a ChatGPT version of the email with those written by students. That might include examining in AI and in human texts formal elements of the genre, language register used, formulaic sequences, and typical rhetorical moves. Having students examine the specific genre characteristics can raise genre awareness so that students can transfer their genre analysis skills to other tasks, particularly those they might encounter in their professional lives (Poole & Polio, 2023).

This kind of rhetorical training is a valuable service that AI can help deliver, provided that it occurs in an instructional context that stresses critical thinking and careful textual analysis. The sample student email generated by ChatGPT provides a good illustration of both the positive and problematic aspects of AI-generated texts. The e-mail contains all the genre-typical formal elements and rhetorical moves and is written in an appropriate register. However, it is overly long and repetitive and obsequious in tone with awkward and inappropriate comments such as "I'm aware of the importance of quizzes as an assessment tool" (Poole & Polio, p. 261). Comparing that text with those generated by students would likely illustrate a greater awareness of the importance of language fitting the social setting and context.

In an L2 instructional setting, in which intercultural awareness is represented by a diversity of student backgrounds or in which developing intercultural competence is an instructional goal, there might be analysis of the email from a multicultural, pragmatic perspective. That might entail discussion of how direct requests might be made (to retake the quiz in this case) or how apologies are formulated (from student for missing the quiz). Such activities can illustrate how genres represent forms of social action (Miller, 1984) and therefore are subject to shaping by the communities from which they arise. Sydorenko et al. (2024) provide a real-world illustration of cultural distinctions in genre characterizations. In analyzing discourse completion tasks by AI, the researchers—coming from US and Hungarian backgrounds—differed in their evaluation of appropriate language registers, with the Hungarian colleague finding the more formal language generated by AI more appropriate than his US colleague. At the same time, we need to be cautious in applying cultural explanations to how speech acts (or genres) play out, avoiding "falling into the trap of presenting L2 learners as people with essentially predictable behaviour" (House & Kádár, 2023, p. 8) and in the process "degrading learners to cultural robots" (p. 8). Instead, speech act realization should be viewed individually from a bottom-up, fine-tuned perspective (House & Kádár, 2023). Caution is all the more called for in working with a language corpus such as that represented in an LLM, which is so clearly culturally skewed.

Understanding the social and cultural identities baked into how genres are used points to the importance of consideration of audience receptivity. That sensitivity to the expectations of a reading or listening audience is problematic for AI: "AI text generators (currently) have a more difficult time simulating the positioned perspectives that human writers bring to situations and communicate to audiences through their genre usage" (Omizo and Hart-Davidson, 2024, p. 272). The training data for AI systems will have included many different genres and their conventions, but AI is devoid of knowledge of situated language use and recipient design. Omizo and Hart-Davidson (2024) maintain that "LLMs do not engage in the full human activity of writing" (p. 272) but rather simulate one aspect of the process through *genre signaling*. Genre signals are specific textual features that have structural and sequential importance in conveying to an audience the use of a socially recognized form of communication, a genre. In AI output the genre form is there, but it is hollow, an empty shell. Given that AI is not capable of effective audience responsiveness in its output, that makes human writing all the more essential: "The capacity to know, empathise with and desire to influence audiences is rapidly emerging as an essential human quality that may future-proof the need for human writers" (McKnight & Gannon, 2024, p. 77). That raises the importance, in teaching writing, of reflecting on the role of audience and in guiding students on how to be effective writers in real-world contexts, within a given human community, whether that be culturally or professionally defined.

**Conclusion**

In writing about AI systems, one should note that they are in a rapid process of change, so that generalizations are problematic. It is also the case that AI systems are likely to improve through user interactions added to their language models, through enlarging their datasets, and through multimodal incorporation (adding video and image training). However, those measures still will not supply the lived experience humans go through in negotiating common ground linguistically and culturally in social interactions and therefore the ability to deal with nuanced pragmatic scenarios. AI generated language–while valuable as a resource in language learning–will remain artificial and inauthentic in ways that cannot serve as an acceptable substitute for actual learner engagement in the L2 with peers and expert speakers. That assertion necessarily leads to the question, what then is the usefulness of AI for language learning and teaching. That is a question that is made even more difficult in that the rapid evolution of AI may render any suggestions quickly out of date.

Spoken dialogue systems have been shown to be helpful for novice level learners to practice pronunciation and to try out their developing L2 skills in dialogic interactions (Bibauw et al., 2022; Godwin-Jones, 2022). With AI chatbots, there is additionally the option of receiving feedback on grammar, word choice, or other L2 features. Those advantages may make it worthwhile for more advanced learners to practice their L2 with AI, especially if they lack conversation partners.

Learners at all levels should be aware, however, of the artificiality of those exchanges and not assume they are gaining the same L2 skills that come from interactions with human partners. With AI, they will not be learning pragmatic behaviors through negotiation of meaning and thus developing interactive and strategic competence. That limitation may change as AI systems are trained with multimodal input and gain capabilities to interpret human tone of voice to gauge users' intent and emotional state. Computer vision, enabled through cameras built into smartphones, VR headsets, or robots, could equip AI with the ability to see and interpret human nonverbal behaviors, an essential element in human meaning-making.

For helping develop L2 writing skills, AI likewise offers both benefits and limitations. AI can assist in many facets of L2 writing: brain-storming ideas, suggesting story lines, and providing feedback of a variety of kinds (grammar, vocabulary, style, structure). Due to its extensive training data, AI systems have knowledge of conventional features for all kinds of writing. That allows them to provide guidance on what might be missing or inappropriate for a specific genre. While AI can supply valuable assistance in formal and structural elements of a text, their textual output will always require critical assessing and correction. Depending on the genre and on the intended audience, minimal adjustments may suffice. In other cases–such as in academic writing settings–substantial reworking will likely be needed. As discussed above, AI output is coherent but uninspiring. It lacks the creativity and personal voice that humans are capable of. AI assistance will be ineffective in guiding writers towards tailoring their texts to the needs and expectation of intended readers. To provide learners with experience in considering texts from the perspective of audience receptivity, it can be advisable to provide writing assignments that have a real-world validity.

Given the linguistic and cultural inauthenticity of AI, L2 learners will need more than reliance on synthetic partners to develop full proficiency, whether that be for oral or written skills. AI only simulates human communication and so needs to be supplemented with interactions with human beings. That could occur in person or in online environments. In instructed settings, virtual exchange provides an ideal vehicle for peer learning. In conversations with fellow learners, particularly in lingua franca environments, there is a need to negotiate not only language forms but sociocultural values and behaviors as well. Pragmatic abilities gained through such interactions are the key to real-world language competence. In addition to virtual exchange, there are many other opportunities online for L2 use in affinity groups or social media. It is important for learners to realize–and for teachers to emphasize–that while AI can provide significant help in a variety of ways to practice and learn another language, ultimately it takes real human interaction in the L2 to develop the ability to use language in way that is socially and culturally appropriate.